\documentclass[10pt,twocolumn,letterpaper]{article}

\usepackage{cvpr}
\usepackage{times}
\usepackage{epsfig}
\usepackage{graphicx}
\usepackage{amsmath}
\usepackage{amssymb}
\usepackage{multirow}
\usepackage{subfigure}
\usepackage{enumerate}
\usepackage{balance}
\usepackage{authblk}
\usepackage{paralist}
\usepackage{epstopdf}

\usepackage[usenames,dvipsnames,svgnames,table]{xcolor}

\usepackage[pagebackref=true,breaklinks=true,letterpaper=true,colorlinks,bookmarks=false]{hyperref}

\cvprfinalcopy 

\def\cvprPaperID{1202} 

\begin{document}

\title{Deep Unsupervised Saliency Detection: A Multiple Noisy Labeling Perspective}

\author[1,2]{Jing Zhang\thanks{These authors contributed equally in this work.}}
\author[2,3]{Tong Zhang$^{\scriptsize*}$}
\author[1]{Yuchao Dai\thanks{Y. Dai (daiyuchao@nwpu.edu.cn) is the corresponding author.}}
\author[2,3]{Mehrtash Harandi}
\author[2]{Richard Hartley}
\affil[1]{Northwestern Polytechnical University, Xi'an, China}
\affil[2]{Australian National University, Canberra, Australia }
\affil[3]{DATA61,CSIRO, Canberra, Australia}

\maketitle

\begin{abstract}

The success of current deep saliency detection methods heavily depends on the availability of large-scale supervision in the form of per-pixel labeling. Such supervision, while labor-intensive and not always possible, tends to hinder the generalization ability of the learned models. By contrast, traditional handcrafted features based unsupervised saliency detection methods, even though have been surpassed by the deep supervised methods, are generally dataset-independent and could be applied in the wild. This raises a natural question that \emph{``Is it possible to learn saliency maps without using labeled data while improving the generalization ability?''}. To this end, we present a novel perspective to unsupervised \footnote{There could be multiple definitions for unsupervised learning, in this paper, we refer unsupervised learning as learning without task-specific human annotations, \eg dense saliency maps in our task.} saliency detection through learning from multiple noisy labeling generated by ``weak'' and ``noisy'' unsupervised handcrafted saliency methods. Our end-to-end deep learning framework for unsupervised saliency detection consists of a latent saliency prediction module and a noise modeling module that work collaboratively and are optimized jointly. Explicit noise modeling enables us to deal with noisy saliency maps in a probabilistic way. Extensive experimental results on various benchmarking datasets show that our model not only outperforms all the unsupervised saliency methods with a large margin but also achieves comparable performance with the recent state-of-the-art supervised deep saliency methods.

\end{abstract}

\section{Introduction}
Saliency detection aims at identifying the visually interesting objects in images that are consistent with human perception, which is intrinsic to various vision tasks such as context-aware image editing \cite{image_editing_2009}, image caption generation \cite{Xu2015show}. Depending on whether human annotations have been used, saliency detection methods can be roughly divided as: unsupervised methods and supervised methods. The former ones compute saliency directly based on various priors (\eg, center prior \cite{ContextSaliency}, global contrast prior \cite{Global-Contrast:CVPR-2011}, background connectivity prior \cite{Background-Detection:CVPR-2014} and \etc), which are summarized and described with human knowledge. The later ones learn direct mapping from color images to saliency maps by exploiting the availability of large-scale human annotated database.


\begin{figure}[t]
\begin{center}
\includegraphics[width=0.95\linewidth]{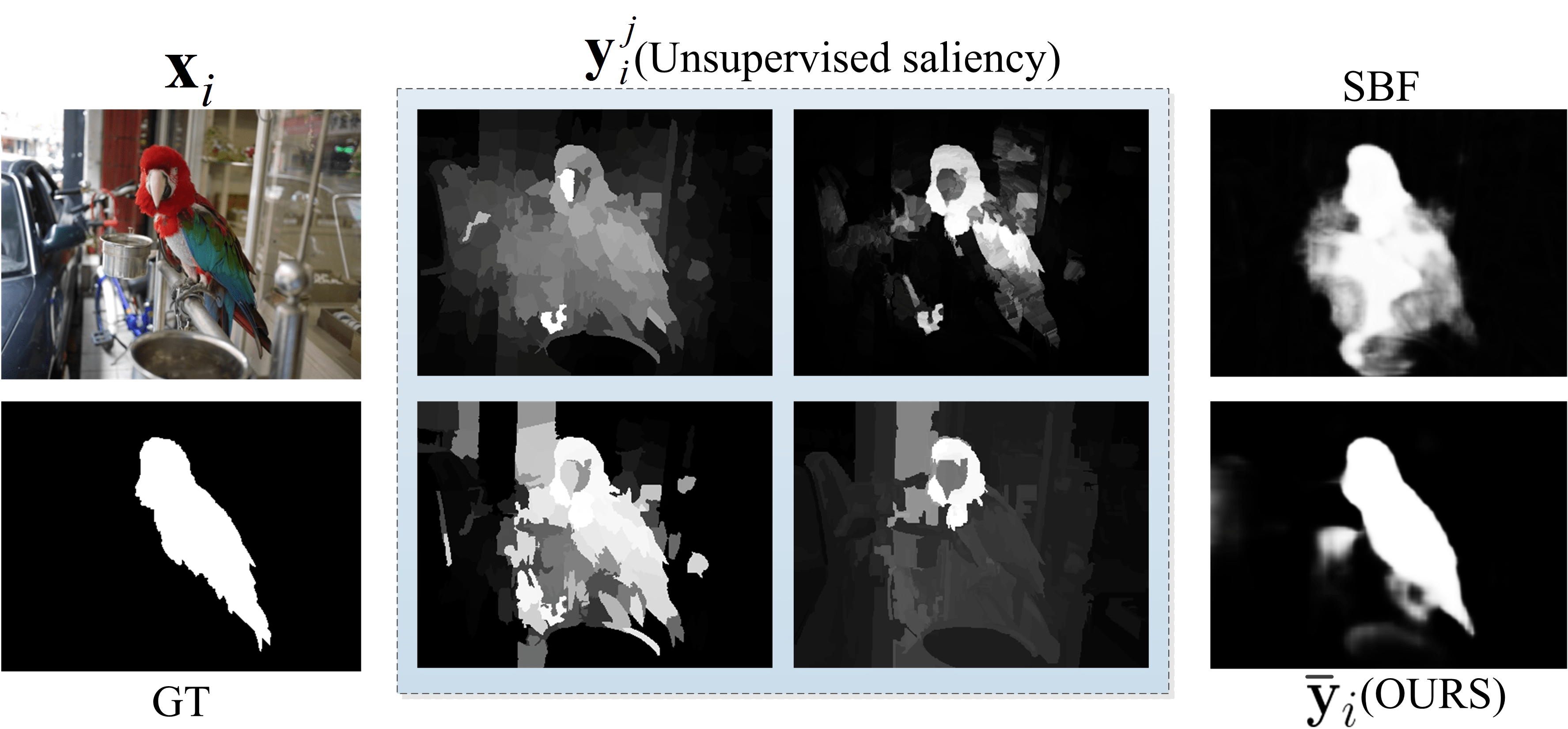}
\caption{\small Unsupervised saliency learning from weak ``noisy'' saliency maps. Given an input image $\mathbf{x}_i$ and its corresponding unsupervised saliency maps $\mathbf{y}_i^j$, our framework learns the latent saliency map $\bar{\mathbf{y}}_i$ by jointly optimizing the saliency prediction module and the noise modeling module. Compared with SBF \cite{Zhang_2017_ICCV} which also learns from unsupervised saliency but with different strategy, our model achieves better performance.}
\label{fig:cmp}
\end{center}
\vspace{-2em}
\end{figure}



Building upon the powerful learning capacity of convolutional neural network (CNN), deep supervised saliency detection methods \cite{DeepMC,ChengCVPR17,Amulet_ICCV} achieve state-of-the-art performances, outperforming the unsupervised methods by a wide margin. The success of these deep saliency methods strongly depend on the availability of large-scale training dataset with pixel-level human annotations, which is not only labor-intensive but also could hinder the generalization ability of the learned network models. By contrast, the unsupervised saliency methods, even though have been outperformed by the deep supervised methods, are generally dataset-independent and could be applied in the wild. 


In this paper, we present a novel end-to-end deep learning framework for saliency detection that is free from human annotations, thus ``unsupervised'' (see Fig.~\ref{fig:cmp} for a visualization). Our framework is built upon existing efficient and effective unsupervised saliency methods and the powerful capacity of deep neural network. The unsupervised saliency methods are formulated with human knowledge and different unsupervised saliency methods exploit different human designed priors for saliency detection. They are noisy (compared with ground truth human annotations) and could have method-specific bias in predicting saliency maps. By utilizing existing unsupervised saliency maps, we are able to remove the need of labor-intensive human annotations, also by jointly learn different priors from multiple unsupervised saliency methods, we are able to get complementary information of those unsupervised saliency.


To effectively leverage these noisy but informative saliency maps, we propose a novel perspective to the problem: \emph{{Instead of removing the noise in saliency labeling from unsupervised saliency methods with different fusion strategies \cite{Zhang_2017_ICCV}, we explicitly model the noise in saliency maps}}. As illustrated in Fig.~\ref{fig:saliency_modules}, our framework consists of two consecutive modules, namely a saliency prediction module that learns the mapping from a color image to the ``latent'' saliency map based on current noise estimation and the noisy saliency maps, and a noise modeling module that fits the noise in noisy saliency maps and updates the noise estimation in different saliency maps based on updated saliency prediction and the noisy saliency maps. In this way, our method takes advantages of both probabilistic methods and deterministic methods, where the latent saliency prediction module works in a deterministic way while the noise modeling module fits the noise distribution in a probabilistic manner.  Experiments suggest that our strategy is very effective and it only takes several rounds \footnote{In our paper, an epoch means a complete pass through all the training data, an iteration means a complete pass through a batch, and a round means an update on noise module.} till convergence.

To the best of our knowledge, the idea of considering unsupervised saliency maps as learning from multiple noisy labels is brand new and different from existing unsupervised deep saliency methods (\eg,~\cite{Zhang_2017_ICCV}). Our main contributions can be summarized as:
\vspace{-2mm}
\begin{enumerate}[1)]
\item We present a novel perspective to unsupervised deep saliency detection, and learn saliency maps from multiple noisy unsupervised saliency methods. We formulate the problem as joint optimization of a latent saliency prediction module and a noise modeling module.
\vspace{-2mm}
\item Our deep saliency model is trained in an end-to-end manner without using any human annotations, leading to an extremely cheap solution.
\vspace{-2mm}
\item Extensive performance evaluation on seven benchmarking datasets show that our framework outperforms existing unsupervised methods with a wide margin while achieving comparable results with state-of-the-art deep supervised saliency detection methods \cite{ChengCVPR17,Amulet_ICCV}.
\end{enumerate}

\section{Related Work}
\label{sec:related_works}
Depending on whether human annotations are used or not, saliency detection techniques can be roughly grouped as unsupervised and supervised methods. Deep learning based methods are particular examples of the latter one. We will also discuss learning with multiple noisy labels.

\subsection{Unsupervised Saliency Detection}
\label{subsec:conventional_methods}
Prior to the deep learning revolution, saliency methods mainly relied on different priors and handcrafted features \cite{Background-Detection:CVPR-2014,Soft-Image-Abstraction:ICCV-2013,Global-Contrast:CVPR-2011,ContextSaliency}. We refer interested readers to \cite{Salient-Detection-Survey:2014} and \cite{SalObjBenchmark_Tip2015} for surveys and benchmark comparisons. Color contrast prior has been exploited at superpixel level in \cite{Global-Contrast:CVPR-2011}. Shen and Wu \cite{Low-Rank-Recovery:CVPR-2012} formulated saliency detection as a low-rank matrix decomposition problem by exploiting the sparsity prior for salient objects. Objectness, which highlights the object-like regions, has also been used in \cite{UFO-Saliency:ICCV-2013} to mark the regions that have higher possibilities of being an object. Zhu \etal ~\cite{Background-Detection:CVPR-2014} presented a robust background measure, namely ``boundary connectivity'' along with an optimization framework to measure backgroundness of each superpixel. Building upon the center prior, \cite{ContextSaliency} detects the image regions that represent the scene, especially those that are near image center.


\subsection{Supervised Saliency Detection}
Conventional supervised techniques, such as \cite{DRFI:CVPR-2013,High-Dim-Color-Transform:CVPR-2014}, formulate saliency detection as a regression problem, and a classifier is trained to assign saliency at pixel or superpixel level. Recently, deep neural networks have been adopted successfully for saliency detection \cite{Amulet_ICCV,NLDF_CVPR,UCF_ICCV,wangiccv17,ChengCVPR17,TIP,DeepMC,MDF:CVPR-2015,RFCN,DC,Jing-Zhang:WACV-2017,Jing-Zhang:ICIP-2017,Jing-Zhang:DICTA-2017}. Deep networks can encode high-level semantic features and hence capture saliency more effectively than both unsupervised saliency methods and non-deep supervised methods. Deep saliency detection methods generally train a deep neural network to assign saliency to each pixel or superpixel. Li and Yu \cite{MDF:CVPR-2015} used learned features from an existing CNN model to replace the handcrafted features. Recently, Cheng \etal \cite{ChengCVPR17} proposed a deep supervised framework with multi-branch short connections embed both high- and low-level features for accurate saliency detection. With the same purpose, a multi-level deep feature aggregation framework is proposed in \cite{Amulet_ICCV}. A top-down strategy and a loss function which penalizes errors on the edge is presented in \cite{NLDF_CVPR}.


\begin{figure*}[!htp]
\begin{center}
\includegraphics[width=0.75\linewidth]{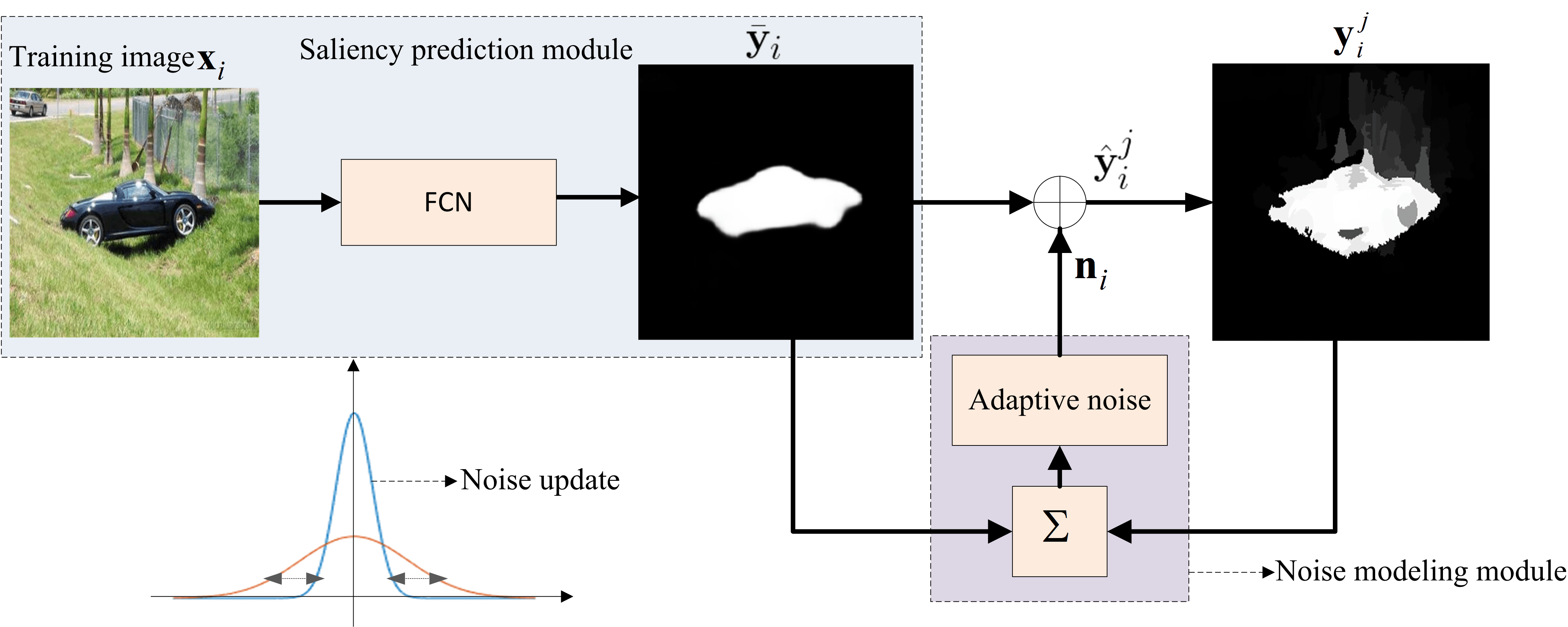}
\caption{\label{fig:saliency_modules}
\small Conceptual illustration of our saliency detection framework, which consists of a ``latent'' saliency prediction module and a noise modeling module. Given an input image, noisy saliency maps are generated by handcrafted feature based unsupervised saliency detection methods. Our framework jointly optimizes both modules under a unified loss function. The saliency prediction module targets at learning latent saliency maps based on current noise estimation and the noisy saliency maps. The noise modeling module updates the noise estimation in different saliency maps based on updated saliency prediction and the noisy saliency maps. In our experiments, the overall optimization converges in several rounds.
}
\end{center}
\end{figure*}
\subsection{Learning with Noisy Labels}
Though deep techniques are methods of choice in saliency detection, very few studies have explicitly addressed the problem of saliency learning with unreliable and noisy labels~\cite{Zhang_2017_ICCV}. Learning with noisy labels is mainly about learning classification models in the presence of inaccurate class labels. Whitehill \etal \cite{NIPS2009_3644} solved the problem of picking the correct label based on the labels provided by many labelers with different expertise. Jindal \etal \cite{ICDM2016} proposed a dropout-regularized noise model by augmenting existing deep network with a noise model that accounts for label noise. Yao \etal \cite{NoisyEmbedding} proposed a quality embedding model to infer the trustworthiness of noisy labels. Different from the above supervised learning with noisy labels methods, Lu \etal \cite{NoisySemantic} proposed a weakly supervised semantic segmentation framework to deal with noisy labels.



To the best of our knowledge, \cite{Zhang_2017_ICCV} is the first and only deep method that learns saliency without human annotations, where saliency maps from unsupervised saliency methods are fused with manually designed rules in combining  ``intra-image'' fusion stream and ``inter-image'' fusion stream to generate the learning curriculum. The method iteratively replaces inter-image saliency map of low reliability with its corresponding saliency map. Their recursive optimization depends on dedicated design and is computationally expensive. Different from \cite{Zhang_2017_ICCV}, we formulate unsupervised saliency learning as the joint optimization of latent saliency and noise modeling. Our method is not only simpler and easier to implement, but also outperforms \cite{Zhang_2017_ICCV} and existing unsupervised saliency methods. Furthermore, our method produces competitive performances as compared to the most recent deep supervised saliency detection methods.

\section{Our Framework}

Targeting at achieving deep saliency detection without human annotations, we propose an end-to-end noise model integrated deep framework, which builds upon existing efficient and effective unsupervised saliency detection methods and the powerful capacity of deep neural networks.


Given a color image $\mathbf{x}_i$, we would like to learn a better saliency map from its $M$ noisy saliency maps $\mathbf{y}_i^j, j=1,\cdots, M$ using different unsupervised saliency methods \cite{Hierarchical:CVPR-2013,MC_ICCV13,DSR_ICCV13,Background-Detection:CVPR-2014}. A trivial and direct solution would be using the noisy saliency maps as ``proxy'' human annotations and train a deep model with these noisy saliency maps as supervision. However, it is well-known that the network training is highly prone to the noise in supervision signals. A simple fusion of the multiple labels (training with averaging, treating as multiple labels) will also not work due to the strong inconsistency between labels. While there could be many other potentials in utilizing the noisy saliency maps, they are all based on human-designed pipelines, thus cannot effectively exploit the best manner. Instead, we propose a principled way to infer the saliency maps from using multiple noisy labels and simultaneously estimate the noise.

\subsection{Joint Saliency Prediction and Noise Modeling}
By contrast to existing manually designed procedures and deep learning based pipeline \cite{Zhang_2017_ICCV}, we propose a new perspective toward the problem of learning from unsupervised saliency. As illustrated in Fig.~\ref{fig:saliency_modules}, our framework consists of two consecutive modules, namely a saliency prediction module that learns the mapping from a color image to the ``latent'' saliency map, and a noise modeling module that fits the noise. These two modules work collaboratively toward fitting the noisy saliency maps. By explicitly modeling noise, we are able to train a deep saliency prediction model without any human annotations and thus achieve unsupervised deep saliency detection.

\subsection{Loss Function}

We start with a set of training images, denoted as $\mathbf{X} = \{\mathbf{x}_i, i = 1, \ldots, N\}$ and a set of $M$ different saliency maps of these images, denoted as $\mathbf{Y} = \{\mathbf{y}_i^j, i = 1,\ldots, N ; j = 1, \ldots,M \}$, where $N$ is number of training images. These are precomputed by applying $M$ different handcrafted ``labellers''. Throughout this discussion, $i$ indexes the training image and $j$ indexes the handcrafted labeller. We propose a neural network with parameter $\Theta$ for saliency detection, which computes a saliency map $\bar{\mathbf{y}}_i = f(\mathbf{x}_i, \Theta)$ of each image. Our idea is to model each of the handcrafted labellers as the sum of $\bar{\mathbf{y}}_i$ plus noise: $\mathbf{y}_i^j = \bar{\mathbf{y}}_i +\mathbf{n}_i^j$, where $\mathbf{n}_i^j$ is a sample chosen from some probability (``noise'') distribution $q_i$, which is to be estimated. For simplicity in this work, it is assumed that the distribution $q$ depends on $\mathbf{x}_i$, and not on the labeller $j$\footnote{Assuming that distribution $q$ is also dependent on the labeller $j$ was observed not to improve results}. We assume a simple model for the noise distributions $q_i$, namely that it a zero-mean Gaussian, independent for each pixel of each image $\mathbf{x}_i$. Thus, the total distribution $ \mathbf{q} = \{ q_1, q_2, \ldots, q_N \} $ is assumed independent for all $i$ and pixel $(m,n)$, and is parametrized by a parameter set $\Sigma = \{ \sigma_{mn}^i \}$, where  $i$ $\,$indexes the training image and $(m,n)$ are pixel coordinates. Sometimes, distribution $\mathbf{q}$ will be denoted as $\mathbf{q}(\Sigma)$ to emphasize the role of the parameters $\Sigma$. With this simple parameterization it is easy to generate noise samples $\mathbf{n}_i^j$ for any $i$ and $j$.

Given $\Theta$, $\Sigma$, and an input image $\mathbf{x}_i$, one generates saliency map $\hat{\mathbf{y}}_i^j$ according to:
\begin{equation}\label{assumtion}
\hat{\mathbf{y}}_i^j = f(\mathbf{x}_i; \Theta) + \mathbf{n}_i^j = \bar{\mathbf{y}}_i +\mathbf{n}_i^j,
\end{equation}
where each $\mathbf{n}_i^j$ is a sample drawn from distribution $q_i(\Sigma)$. In the training process, the parameters $\Theta$ of the network and $\Sigma$ of the noise model are updated to minimize an appropriate loss function. The loss function has two parts:
\begin{equation}\label{cost}
\mathcal{L}(\Theta,\Sigma) = \mathcal{L}_{\rm pred}(\Theta,\Sigma) + \lambda \mathcal{L}_{\rm noise}(\Theta,\Sigma),
\end{equation}
where $\lambda$ is the regularizer to balance these two terms. Under our optimization framework, increasing the variance in noise modeling will make the prediction loss $\mathcal{L}_{\rm pred}$ large and decrease the $\mathcal{L}_{\rm noise}$. Meanwhile, keeping the variance lower will decrease the cross-entropy loss $\mathcal{L}_{\rm pred}$ but increase $\mathcal{L}_{\rm noise}$. Thus our model balances between these two losses and converges to the state minimizing the overall loss.
These two losses are described below:
\paragraph{Saliency Prediction:} For the latent saliency prediction module, we use a fully convolutional neural network (FCN) due to its superior capability in feature learning and feature representation. We use the conventional cross-entropy loss and compute the loss function element-wisely across the whole training images.

The predictive loss $\mathcal{L}_{\rm Pred}$ is designed to measure the agreement of the predicted labellings $\hat{\mathbf{y}}_i^j$ with handcrafted labellings $\mathbf{y}_i^j$. Cross-entropy loss is used for this purpose, and the cross-entropy loss between modeled value $\hat{y}$ and ``ground truth'' value $y$ (noisy label) is given by:
\begin{equation}
L_{\rm CE} = -( y\log(\hat{y}) + (1-y)\log(1-\hat{y})).
\end{equation}
This is applied to all pixel $(m,n)$, all labellers $j$ and all the test images $\mathbf{x}_i$ to give the total prediction loss.

\begin{equation}\label{cost_pred}
\begin{split}
\mathcal{L}_{\rm pred}(\Theta,\mathbf{\Sigma}) = & \sum_{i =1}^N\sum_{j =1}^M\sum_{m,n}L_{\rm CE}(\mathbf{y}_{i,mn}^j,\hat{\mathbf{y}}^j_{i,mn}),
\end{split}
\end{equation}
where $\hat{\mathbf{y}}_{i,mn}^j$ is our noisy saliency map prediction at pixel $(m,n)$ which can be easily computed by~\eqref{assumtion} element-wise, and $\hat{\mathbf{y}}_{i,mn}^j$ is truncated to lie in the range of $[0, 1]$.


\paragraph{Noise Modeling}

To effectively handle noisy saliency maps from different unsupervised saliency map labelers, we build a probabilistic model to approximate the noise, and connect it with our deterministic part (latent saliency prediction model as shown in Fig.~\ref{fig:saliency_modules}). In this way, our entire model can be trained in an end-to-end manner to minimize the overall loss function Eq.~\eqref{cost}.

The noise loss $\mathcal{L}_{\rm noise}$ measures (for each training image $\mathbf{x}_i$) the agreement of the noise distribution $q_i(\mathbf{\Sigma})$ with the empirical variance of the measurements $\mathbf{y}_i^j$ with respect to the output $\bar{\mathbf{y}}_i = f(\mathbf{x}_i;\Theta)$ of the network.
More precisely, given an input $\mathbf{x}_i$, define $\hat{\mathbf{n}}_i^j = \mathbf{y}_i^j - \mathbf{\bar{y}}_i$, the empirical error of each $\mathbf{y}_i^j$ with respect to the network prediction. For each pixel location $(m,n)$, this provides $M$ samples from a zero-mean Gaussian probability distribution $p_i$, and its variance on every pixel can be written as $\hat{\sigma}_{i,mn}$. The complete set of parameters for $p_i$ is denoted as $\hat{\mathbf{\Sigma}}=  \{ \hat{\sigma}_{i,mn} \}$.

Since it is intractable to estimate the true posterior distribution of $\hat{\mathbf{n}}_i^j$, thus we propose to approximate it by sequentially optimizing the parameters of prior. We assume that the noise is generated by some random process, involving an unobserved continuous random variable set $\mathbf{\Sigma}$. From an encoder perspective, the unobserved variable $\mathbf{n}$ can be interpreted as a latent representation. Here, we model $\hat{\mathbf{y}}_i^j$ as a probabilistic encoder, since given an image $\mathbf{x}_i$ and network parameters $\Theta$ it produces a distribution (\eg a Gaussian) over possible values of the code $\mathbf{n}$. The process consists of two steps: (1) a noise map $\mathbf{n}_i$ is generated from some prior distribution $q(\mathbf{\Sigma}^*)$; (2) a noise map $\hat{\mathbf{n}}_i^j$ is produced and estimating the corresponding parameter $\hat{\sigma}_i$

The corresponding noise loss is defined to be the KL divergence between distribution $p_i$ and $q_i$.
\begin{equation}
\begin{split}
\mathcal{L}_{\rm noise}(\Theta,\Sigma) = \sum_i^N \mathbf{KL}(q(\mathbf{\Sigma}_i) \| \, p(\hat{\mathbf{\Sigma}}_i)).
\end{split}
\end{equation}

Since we employ the Gaussian distribution as the prior distribution for our noise model, the KL divergence has a closed-form solution as:
\begin{equation}\label{KL}
\mathbf{KL}(q(\mathbf{\sigma}) \| \, p(\hat{\mathbf{\sigma}})) = \log(\hat{\sigma}/\sigma) + \frac{\sigma^2+(\mu - \hat{\mu})^2}{2\hat{\sigma}^2} - \frac{1}{2},
\end{equation}
Based on this equation, we can update $\sigma^2_{i}$ for every coordinate $(m,n)$ as
\begin{equation}
(\sigma^{t+1}_{i})^2 = (\sigma_{i}^t)^2 + \alpha((\hat{\sigma}^t_{i})^2 - (\sigma_{i}^t)^2),
\end{equation}
by differentiating Eq.~\eqref{KL} with respect to $\sigma_{i,mn}^2$, where $\alpha$ is the step size, and we set $\alpha = 0.01$ in this paper.

For different images we have the corresponding noise maps, which follows i.i.d. Gaussian distributions with different variance. Thus, it is hard to converge if simultaneously optimizing the FCN parameters $\Theta$ and noise parameters $\Sigma$. In order to train the whole network smoothly, we update the parameters of noise module after the prediction loss converges. Noise maps of a given image are sampled from the same distribution in a round, but they are updated in every round. At the first round, we initialize noise variance to be zero, and train the FCN until it converges. Based on the variance of the saliency prediction and noisy labels, we then update the noise variance for each image and retrain the network. Through minimizing the loss function Eq.~\eqref{cost} with this procedure,  We can train the network and estimate the corresponding noise maps.



\subsection{Deep Noise Model based Saliency Detector}
\textbf{Network Architecture} We build our latent saliency prediction module upon the DeepLab network \cite{Deeplab}, where a deep CNN (ResNet-101 \cite{ResHe2015} in particular) originally designed for image classification is re-purposed by 1) transforming all fully connected layers to convolutional layers and 2) increasing feature resolution through dilated convolution \cite{Deeplab}. Figure~\ref{fig:saliency_modules} shows the whole structure of our framework. Specifically, our model takes a rescaled image $\mathbf{x_i}$ of $425 \times 425$ as input. For training, the noise model is used to iteratively update saliency prediction $\hat{\mathbf{y}}_i^j$, and it's excluded in testing stage, where the latent saliency prediction output $\bar{\mathbf{y}}_i$ in Fig. \ref{fig:saliency_modules} is our predicted saliency map.

\textbf{Implementation details:} We trained our model using Caffe \cite{jia2014caffe} with maximum epoch of 20. We initialized our model by using the Deep Residual Model trained for image classification \cite{ResHe2015}. We used the stochastic gradient descent method with momentum 0.9 and decreased learning rate 90\% when the training loss did not decrease. Base learning rate is initialized as 1e-3 with the ``poly'' decay policy \cite{jia2014caffe}. For validation, we set ``test\_iter'' as 500 (test batch size 1) to cover the full 500 validation images. The training took 4 hours for one round with training batch size 1 and ``iter\_size'' 20 on a PC with an NVIDIA Quadro M4000 GPU.

\section{Experimental Results}
In this section, we report experimental results on various saliency detection benchmarking datasets.

\begin{table*}[!htp] \footnotesize
\begin{center}
\caption{\small Performance of mean F-measure ($F_\beta$) and MAE for different methods including ours on seven benchmark datasets.} \centering
\begin{tabular}{l c c c c c c c c c c c c c c }\hline
 & \multicolumn{2}{c}{MSRA-B}  & \multicolumn{2}{c}{ECSSD} & \multicolumn{2}{c}{DUT}  & \multicolumn{2}{c}{SED2} & \multicolumn{2}{c}{PASCAL\-S}   & \multicolumn{2}{c}{THUR} & \multicolumn{2}{c}{SOD}  \\
Methods & $F_\beta$          & MAE         & $F_\beta$          & MAE  & $F_\beta$          & MAE         & $F_\beta$          & MAE   & $F_\beta$          & MAE         & $F_\beta$          & MAE & $F_\beta$          & MAE     \\ \hline
BL1&.7905&.0936&.7205&.1444&.5825&.1369&.7773&.1112&.6714&.2206&.5953&.1339&.6306&.1870   \\
BL2&.6909&.1710&.6542&.2170&.4552&.2951&.7232&.1406&.6776&.2409&.5119&.2545&.5928&.2566   \\

BL3 &.8879&.0587&.8717&.0772&.7253&.0772&.8520&.0819&.8264&.1525&.7368&.0749&.7922&.1231   \\
OURS &.8770&.0560&.8783&.0704&.7156&.0860&.8380&.0881&.8422&.1391&.7322&.0811&.7976&.1182   \\ \hline
\end{tabular}
\label{tab:baseline}
\end{center}
\end{table*}
\begin{table*}[!htp] \footnotesize
\begin{center}
\caption{Performance of mean F-measure ($F_\beta$) and MAE for different methods including ours on seven benchmark datasets (Best ones in bold). From DSS to DC are deep learning based supervised methods, from DRFI to HS are the handcrafted feature based unsupervised methods, SBF and OURS are deep learning based unsupervised saliency detection methods.} \centering
\begin{tabular}{l c c c c c c c c c c c c c c }\hline
 & \multicolumn{2}{c}{MSRA-B}  & \multicolumn{2}{c}{ECSSD} & \multicolumn{2}{c}{DUT}  & \multicolumn{2}{c}{SED2} & \multicolumn{2}{c}{PASCAL\-S}   & \multicolumn{2}{c}{THUR} & \multicolumn{2}{c}{SOD}  \\
Methods & $F_\beta$          & MAE         & $F_\beta$          & MAE  & $F_\beta$          & MAE         & $F_\beta$          & MAE   & $F_\beta$          & MAE         & $F_\beta$          & MAE & $F_\beta$          & MAE     \\ \hline
DSS \cite{ChengCVPR17}&.8941&.0474&.8796&.0699&.7290&\textbf{.0760}&.8236&.1014&.8243&.1546&.7081&.1142&.8048&.1118   \\
NLDF \cite{NLDF_CVPR} &\textbf{.8970}&.0478&\textbf{.8908}&.0655&\textbf{.7360}&.0796&-&-&.8391&.1454&-&-&\textbf{.8235}&\textbf{.1030}   \\
Amulet \cite{Amulet_ICCV}&-&-&.8825&\textbf{.0607}&.6932&.0976&\textbf{.8745}&\textbf{.0629}&.8371&\textbf{.1292}&.7115&.0937&.7729&.1248   \\
UCF \cite{UCF_ICCV}&-&-&.8521&.0797&.6595&.1321&.8444&.0742&.8060&.1492&.6920&.1119&.7429&.1527   \\
SRM \cite{wangiccv17}&.8506&.0665&.8260&.0922&.6722&.0846&.7447&.1164&.7766&.1696&.6894&.0871&.7246&.1369   \\
DMT \cite{TIP}&-&-&.7589&.1601&.6045&.0758&.7778&.1074&.6657&.2103&.6254&.0854&.6978&.1503   \\
RFCN \cite{RFCN}&-&-&.8426&.0973&.6918&.0945&.7616&.1140&.8064&.1662&.7062&.1003&.7531&.1394   \\
DeepMC \cite{DeepMC}&.8966&.0491&.8061&.1019&.6715&.0885&.7660&.1162&.7327&.1928&.6549&.1025&.6862&.1557   \\
MDF \cite{MDF:CVPR-2015}&.7780&.1040&.8097&.1081&.6768&.0916&.7658&.1171&.7425&.2069&.6670&.1029&.6377&.1669   \\
DC \cite{DC}&.8973&\textbf{.0467}&.8315&.0906&.6902&.0971&.7840&.1014&.7861&.1614&.6940&.0959&.7603&.1208   \\
\hline
DRFI \cite{DRFI:CVPR-2013}&.7282&.1229&.6440&.1719&.5525&.1496&.7252&.1373&.5745&.2556&.5613&.1471&.5440&.2046   \\
RBD \cite{Background-Detection:CVPR-2014}&.7508&.1171&.6518&.1832&.5100&.2011&.7939&.1096&.6581&.2418&.5221&.1936&.5927&.2181   \\
DSR \cite{DSR_ICCV13}&.7227&.1207&.6387&.1742&.5583&.1374&.7053&.1452&.5785&.2600&.5498&.1408&.5500&.2133   \\
MC \cite{MC_ICCV13}&.7165&.1441&.6114&.2037&.5289&.1863&.6619&.1848&.5742&.2719&.5149&.1838&.5332&.2435   \\
HS \cite{HierarchySaliency15}&.7129&.1609&.6234&.2283&.5205&.2274&.7168&.1869&.5948&.2860&.5157&.2178&.5383&.2729   \\ \hline
SBF \cite{Zhang_2017_ICCV} &-&-&.7870&.0850&.5830&.1350&-&-&.7780&.1669&-&-&.6760&.1400  \\
OURS &.8770&.0560&.8783&.0704&.7156&.0860&.8380&.0881&\textbf{.8422}&.1391&\textbf{.7322}&\textbf{.0811}&.7976&.1182   \\ \hline
\end{tabular}
\label{tab:deep_unsuper_Performance_Comparison}
\end{center}
\end{table*}

\begin{figure*}[!htp]
   \begin{center}
   {\includegraphics[width=0.31\linewidth]{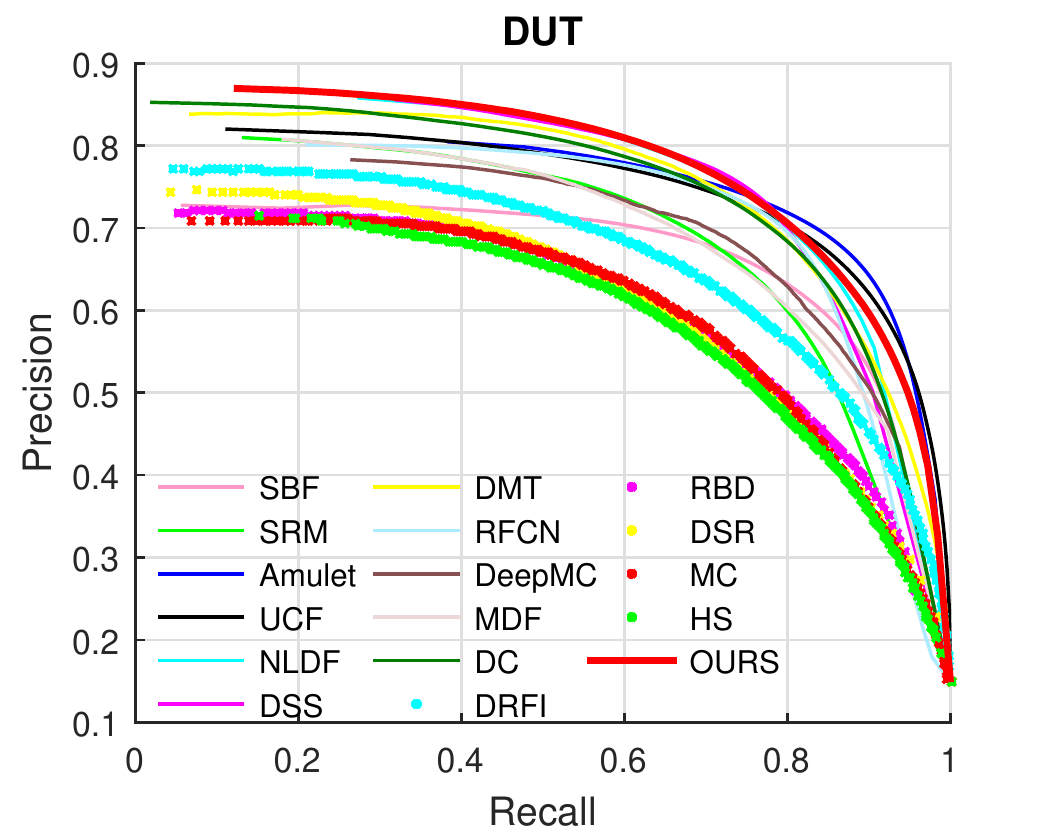}}
   {\includegraphics[width=0.31\linewidth]{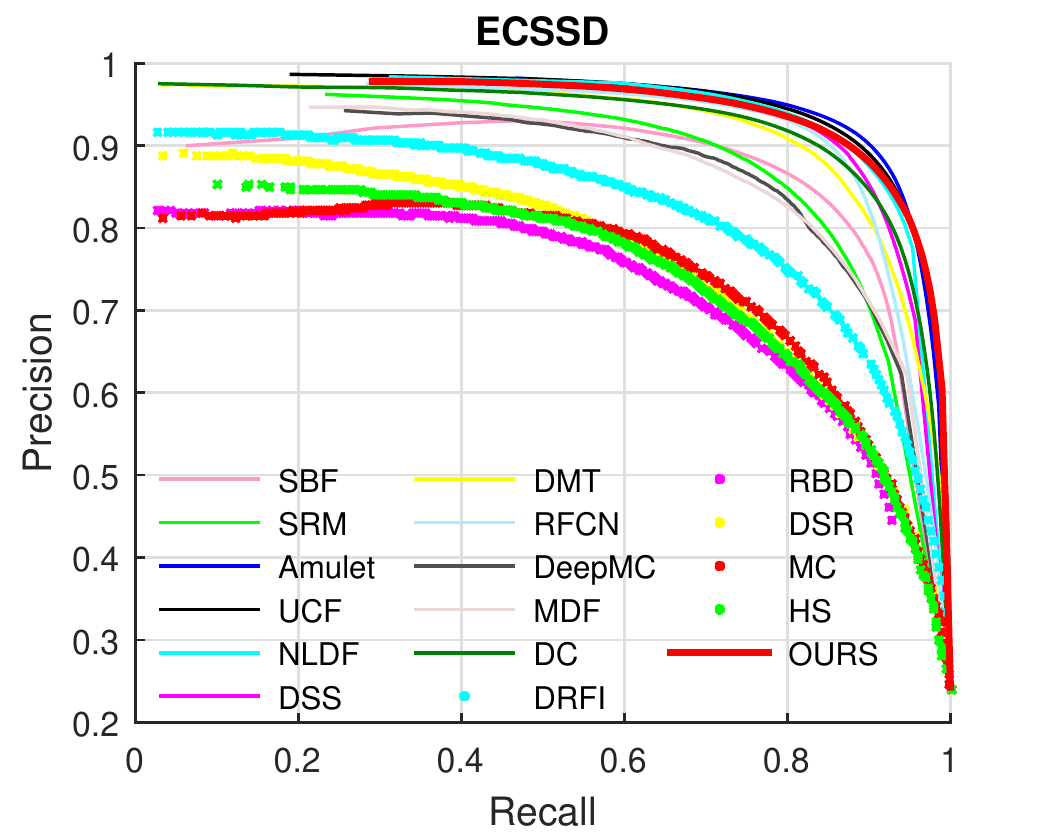}}
   {\includegraphics[width=0.31\linewidth]{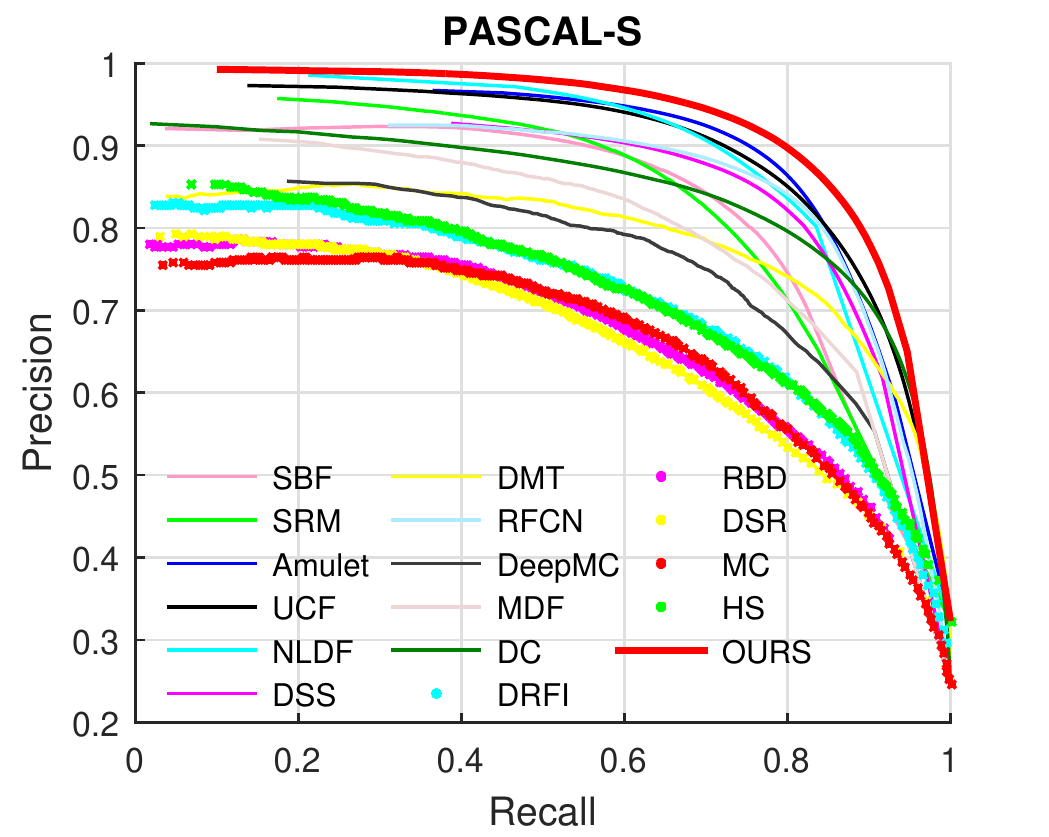}}
   {\includegraphics[width=0.31\linewidth]{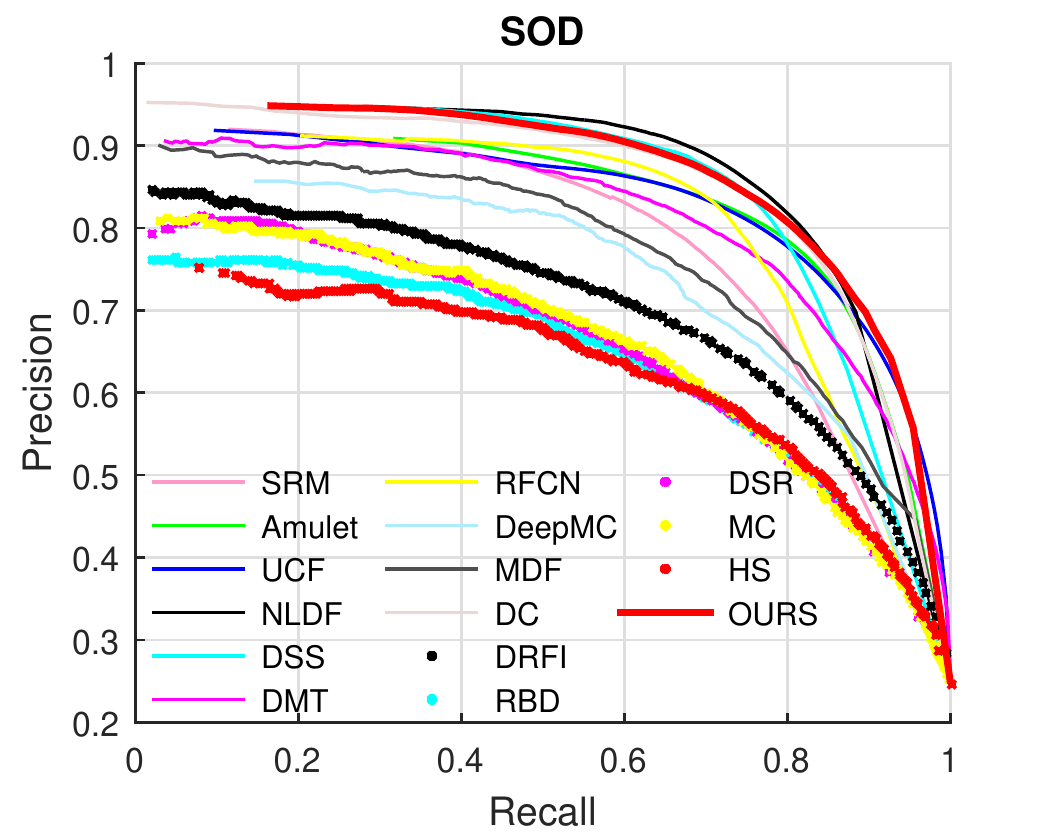}}
   {\includegraphics[width=0.31\linewidth]{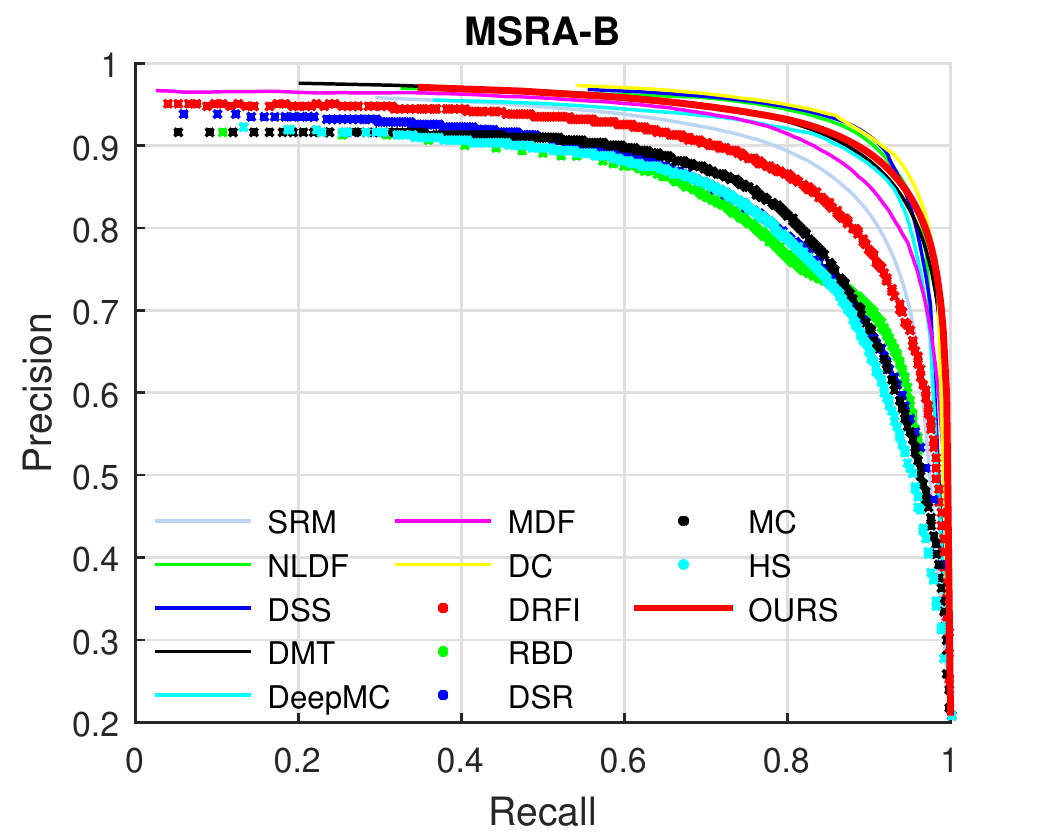}}
   {\includegraphics[width=0.31\linewidth]{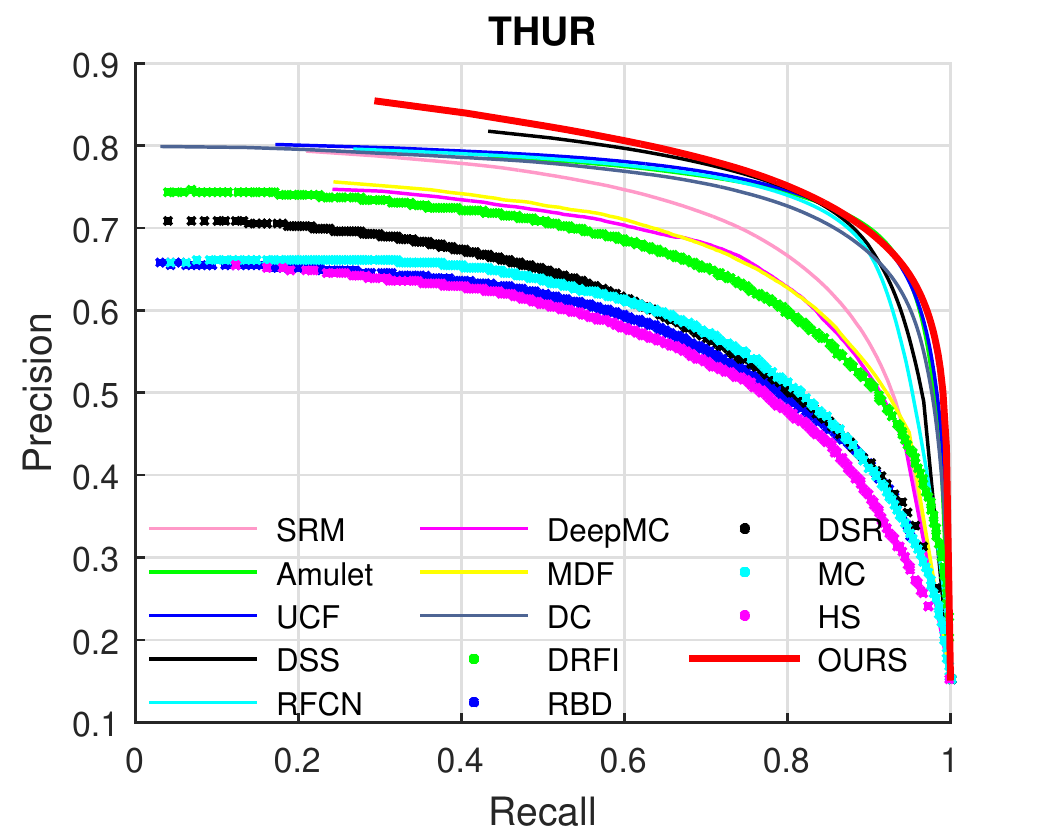}}
   \end{center}
   \vspace{-5mm}
   \caption{PR curves on six benchmark datasets (DUT, ECSSD, PASCAL-S, SOD, MSRA-B, THUR). Best Viewed on Screen.}
   \label{fig:pr_curve}
\end{figure*}
\subsection{Setup}
\label{subsec:experimental_setup}
\textbf{Dataset:} We evaluated performance of our proposed model on 7 saliency benchmarking datasets. 3,000 images from the MSRA-B dataset\cite{Learning-Detect-Salient:CVPR-2007} are used to get the noisy labels (where 2,500 images for training and 500 images for validation) and the remaining 2,000 images are kept for testing. Most of the images in MSRA-B dataset only have one salient object. The ECSSD dataset \cite{Hierarchical:CVPR-2013} contains 1,000 images of semantically meaningful but structurally complex images. The DUT dataset \cite{Manifold-Ranking:CVPR-2013} contains 5,168 images. The SOD saliency dataset \cite{DRFI:CVPR-2013} contains 300 images, where many images contain multiple salient objects with low contrast. The SED2 \cite{SED} dataset contains 100 images with each image contains two salient objects. The PASCAL-S \cite{PASCALS} dataset is generated from the PASCAL VOC \cite{PASCAL_VOC} dataset and contains 850 images. The THUR dataset \cite{THUR} contains 6,232 images of five classes, namely ``butterfly'',``coffee mug'',``dog jump'',``giraffe'' and ``plane''.

\textbf{Unsupervised Saliency Methods:} In this paper, we learn unsupervised saliency from existing unsupervised saliency detection methods. In our experiment, we choose RBD \cite{Background-Detection:CVPR-2014}, DSR \cite{DSR_ICCV13}, MC \cite{MC_ICCV13} and HS \cite{Hierarchical:CVPR-2013} due to their effectiveness and efficiency as illustrated in \cite{SalObjBenchmark_Tip2015}.

\textbf{Competing methods:} We compared our method against 10 state-of-the-art deep saliency detection methods (with clean labels): DSS \cite{ChengCVPR17}, NLDF \cite{NLDF_CVPR}, Amulet \cite{Amulet_ICCV}, UCF \cite{UCF_ICCV}, SRM \cite{Zhang_2017_ICCV}, DMT \cite{TIP}, RFCN \cite{RFCN}, DeepMC \cite{DeepMC}, MDF \cite{MDF:CVPR-2015} and DC \cite{DC}, 5 conventional handcrafted feature based saliency detection methods: DRFI \cite{DRFI:CVPR-2013}, RBD \cite{Background-Detection:CVPR-2014}, DSR \cite{DSR_ICCV13}, MC \cite{MC_ICCV13}, and HS \cite{Hierarchical:CVPR-2013}, which were proven in \cite{SalObjBenchmark_Tip2015} as the state-of-the-art methods before the deep learning revolution, and the very recent unsupervised deep saliency detection method SBF \cite{Zhang_2017_ICCV}.


\textbf{Evaluation metrics:} We use 3 evaluation metrics, including the mean absolute error (MAE), F-measure, as well as the Precision-Recall (PR) curve. MAE can provide a better estimate of the dissimilarity between the estimated and ground truth saliency map. It is the average per-pixel difference between the ground truth and the estimated saliency map, normalized to [0, 1], which is defined as:
\begin{equation}
MAE = \frac{1}{W\times H} \sum_{x=1}^W \sum_{y=1}^H |S(x,y) - GT(x,y)|,
\label{eq:MAR}
\end{equation}
where $W$ and $H$ are the width and height of the respective saliency map $S$, $GT$ is the ground truth saliency map.

The F-measure ($F_{\beta}$) is defined as the weighted harmonic mean of precision and recall:
\begin{equation}
F_{\beta} = (1+\beta^2)\frac{Precision\times Recall}{\beta^2 Precision + Recall},
\label{eq:F_Measure}
\end{equation}
where $\beta^2=0.3$, $Precision$ corresponds to the percentage of salient pixels being correctly detected, $Recall$ is the fraction of detected salient pixels in relation to the ground truth number of salient pixels. The PR curves are obtained by thresholding the saliency map in the range of [0, 255].

\subsection{Baseline Experiments}
As there could be different ways to utilize the multiple noisy saliency maps, and for fair comparisons with straightforward solutions for our task, we run the following three baseline methods and the results are reported in Table~\ref{tab:baseline}.

\textbf{Baseline 1£º using noisy unsupervised saliency pseudo ground truth:} For a given input image $\mathbf{x}_i$ and its $M$ handcrafted feature based saliency map $\mathbf{y}_i^j, j = 1,...,M$, we get $M$ image pairs with noisy label \{$\mathbf{x}_i$,$\mathbf{y}_i^j, j = 1,...,M$). Then we train a deep model \cite{ResHe2015} based on those noisy labels directly, and the results are shown as ``BL1'' in Table~\ref{tab:baseline}.

\textbf{Baseline 2: using averaged unsupervised saliency as pseudo ground truth:} Instead of using all the four unsupervised saliency as ground truth, we use the averaged saliency map of those unsupervised saliency as pseudo ground truth, and trained another baseline model ``BL2'' in Table~\ref{tab:baseline}.


\textbf{Baseline 3: supervised learning with ground truth supervision:} Our proposed framework consists of the saliency prediction module and the noise modeling module to effectively leverage the noisy saliency maps. To illustrate the best performance our model can achieve as well as to provide a baseline comparison for our framework, we train our latent saliency module directly with clean labels, which naturally gives an upper bound of the saliency detection performance. The results ``BL3'' are reported in Table~\ref{tab:baseline}.

\textbf{Analysis:} In Table~\ref{tab:baseline}, we compare our unsupervised saliency method with the above baseline configurations. Our method clearly outperforms both BL1 and BL2 with a wide margin, demonstrating the superiority of our end-to-end learning framework. As illustrated in Table~\ref{tab:baseline}, the performance of BL1 is better than the performance of BL2. This is because: 1) For BL1, we have 12,000 training image pairs (four unsupervised saliency methods), while for BL2, we have 3,000 averaged noisy labels; 2) as those unsupervised saliency methods tend to prefer different priors for saliency detection, and their saliency maps can be complementary or controversial to some extent. Simply averaging those saliency maps results in even worse proxy saliency map supervision. Compared with BL3, which is trained with ground truth clean labels and without noise, our unsupervised method achieves highly comparable results. This demonstrates that by jointly learning the latent saliency maps and modeling the noise in a unified framework, we are able to learn the desired reliable saliency maps even without any human annotations.

\begin{figure*}[!htp]
\centering
\includegraphics[scale=0.43]{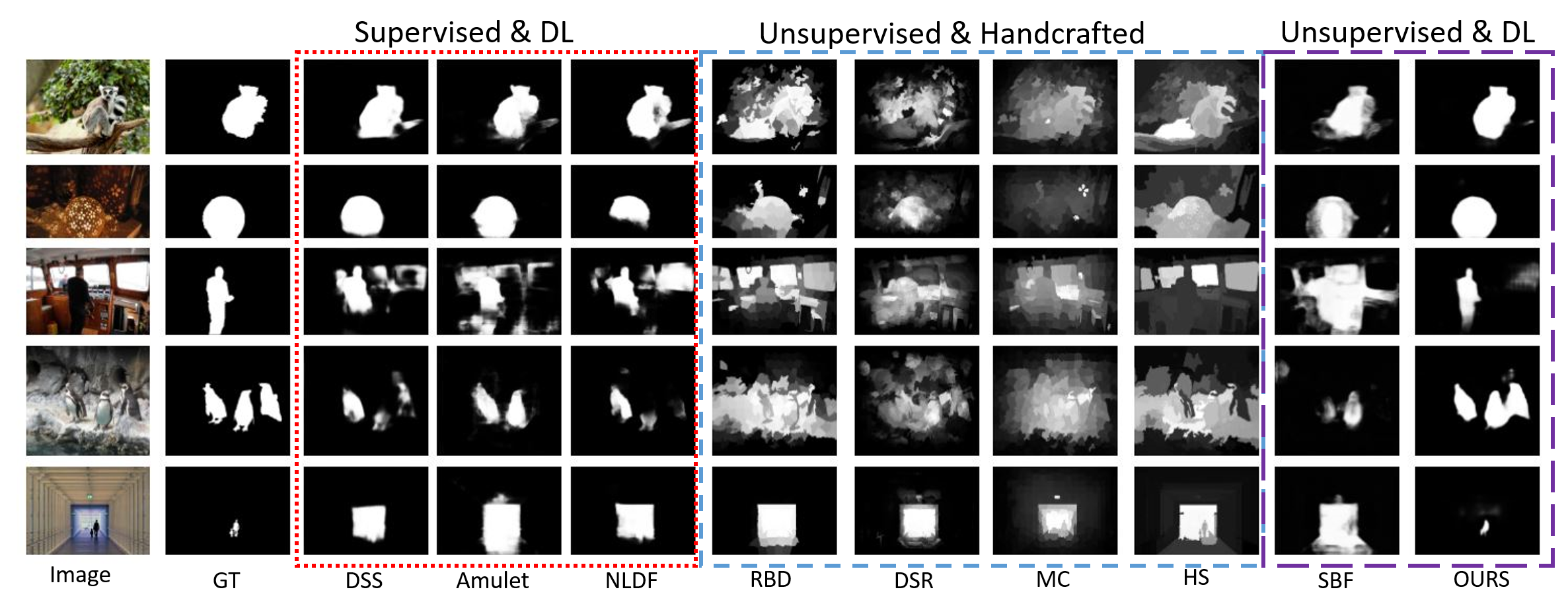}
\caption{Visual comparison between our method and other competing methods.}
\label{fig:sample_show}
\end{figure*}

\subsection{Comparison with the State-of-the-art}
\textbf{Quantitative Comparison}
We compared our method with eleven most resent deep saliency methods and five conventional methods. Results are reported in Table~\ref{tab:deep_unsuper_Performance_Comparison} and Fig.~\ref{fig:pr_curve}, where ``OURS'' represents the results of our model. Table~\ref{tab:deep_unsuper_Performance_Comparison} shows that on those seven benchmark datasets, deep supervised methods significantly outperform traditional methods with 2\%-12\% decrease in MAE, which further proves the superiority of deep saliency detection.

MSRA-B is a relatively simple dataset, where most salient objects dominate the whole image. The most recent deep supervised saliency methods \cite{ChengCVPR17} \cite{NLDF_CVPR} \cite{Amulet_ICCV} can achieve the highest mean F-measure of 0.8970, and our unsupervised method without human annotations can achieve a mean F-measure of 0.8770, which is only a slight worse. The DUT dataset has more than 25\% of images with saliency occupation less than 4\%. Small salient object detection is quite challenging which increase the difficulty of this dataset. We achieve the third highest mean F-measure compared with all the competing methods. The THUR dataset is the largest dataset we used in this paper, and most of the images have complex background. The state-of-the-art competing method achieves a mean F-measure/MAE as 0.7115/0.0854, while our method achieves the best mean F-measure and MAE as 0.7322/0.0811. SBF \cite{Zhang_2017_ICCV} uses inter- and intra-image confidence map as pseudo ground truth to train an unsupervised deep model based on unsupervised saliency, which is quite different from our formulation of predicting saliency from unsupervised saliency as learning from noisy labels. Table ~\ref{tab:deep_unsuper_Performance_Comparison} shows that our framework leads to better performance, with 10\% mean F-measure improvement and 3\% decrease of MAE on average. Fig.~\ref{fig:pr_curve} shows comparison between PR curves of our method and the competing methods on four benchmarking datasets. For the PASCAL-S and THUR dataset, our method ranks almost the 1st, and for the other three datasets, our method achieves competitive performance compared with the competing deep supervised methods. These experiments altogether proves the effectiveness our proposed unsupervised saliency detection framework.

\textbf{Qualitative Comparison}
Figure~\ref{fig:sample_show} demonstrates several visual comparisons, where our method consistently outperforms the competing methods, especially those four unsupervised saliency we used to train our model. The first image is a simple scenario, and most of the competing methods can achieve good results, while our method achieves the best result with most of the background region suppressed. Background of the third image is very complex, and all the competing methods fail to detect salient object. With proper noisy labels, we achieve the best results compared with both unsupervised saliency methods and deep saliency methods. The fourth image is in very low-contrast, where most of the competing methods failed to capture the whole salient objects with the last penguin mis-detected, especially for those unsupervised saliency methods. Our method captures all the three penguins properly. The salient objects in the last row are quite small, and the competing methods failed to capture salient regions, while our method capture the whole salient region with high precision.

\begin{figure}[t]
\begin{center}
\vspace{-2mm}
\subfigure[MAE of each round on 7 datasets]{{\includegraphics[width=1\linewidth]{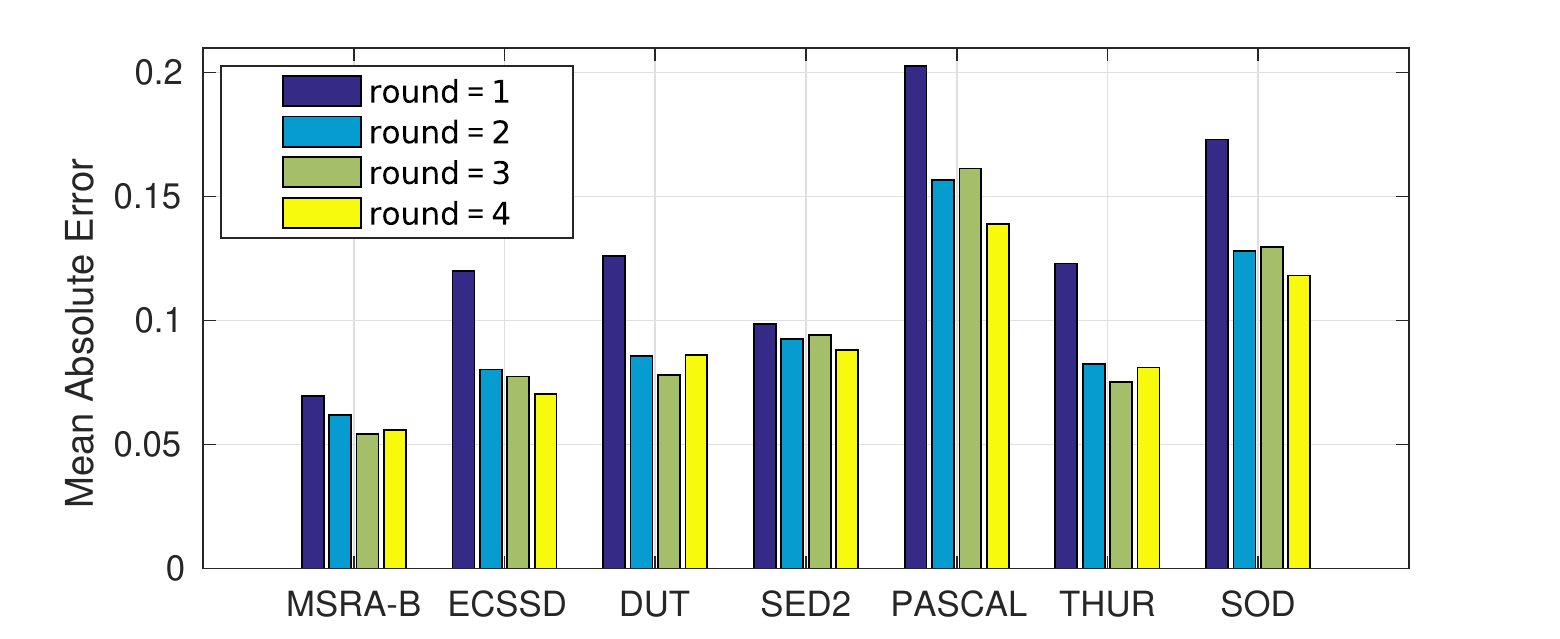}}}
\vspace{-2mm}
\subfigure[Input]{\includegraphics[width=0.30\linewidth]{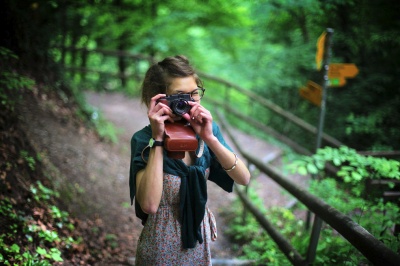}}
\subfigure[GT]{\includegraphics[width=0.30\linewidth]{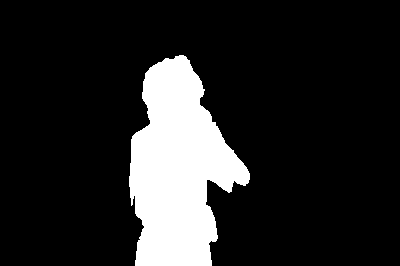}}
\subfigure[1st]{\includegraphics[width=0.30\linewidth]{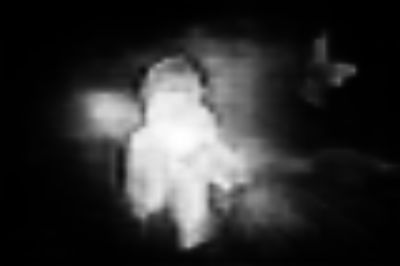}}
\subfigure[2nd]{\includegraphics[width=0.30\linewidth]{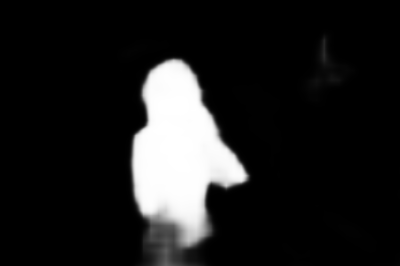}}
\subfigure[3rd]{\includegraphics[width=0.30\linewidth]{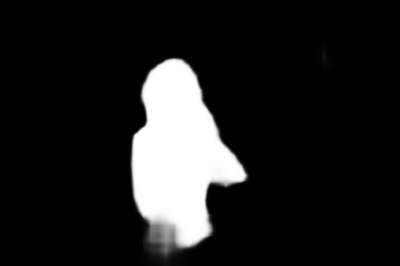}}
\subfigure[4th]{\includegraphics[width=0.30\linewidth]{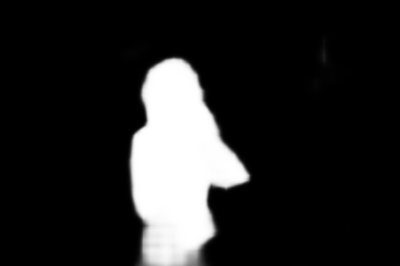}}
\caption{\small Performance of each round. Top: MAE of each dataset. Bottom: an example image, ground-truth and intermedia results generated by each updating round.}
\label{fig:ablation}
\end{center}
\vspace{-2em}
\end{figure}

\textbf{Ablation Studies:} In this paper, we propose to iteratively update the noise modeling module and the latent saliency prediction model to achieve accurate saliency detection. As the two modules work collaboratively to optimize the overall loss function, it is interesting to see how the saliency prediction results evolves with respect to the increase of updating round. In Fig.~\ref{fig:ablation}, we illustrate both the performance metric (MAE) with respect to updating round and an example saliency detection results. Starting with the zero noise initialization, our method consistently improves the performance of saliency detection with the updating of noise modeling. Also, only after several updating rounds, our method convergences to desired state as shown in Fig.~\ref{fig:ablation}.

\section{Conclusions}
In this paper, we propose an end-to-end saliency learning framework without the need of human annotated saliency maps in network training. We represent unsupervised saliency learning as learning from multiple noisy saliency maps generated by various efficient and effective conventional unsupervised saliency detection methods. Our framework consists of a latent saliency prediction module and an explicit noise modeling models, which work collaboratively. Extensive experimental results on various benchmarking datasets prove the superiority of our method, which not only outperforms traditional unsupervised methods with a wide margin but also achieves highly comparable performance with current state-of-the-art deep supervised saliency detection methods. In the future, we plan to investigate the challenging scenarios of multiple saliency object detection and small salient object detection under our framework. Extending our framework to dense prediction tasks such as semantic segmentation \cite{NoisySemantic} and monocular depth estimation \cite{Libo-Depth:CVPR-2015} could be interesting directions.

\noindent\textbf{Acknowledgement.}{\small ~ J. Zhang would like to thank Prof. Mingyi He for his immeasurable support and encouragement. T. Zhang was supported by the Australian Research Council (ARC) Discovery Projects funding scheme (project DP150104645). Y. Dai was supported in part by National 1000 Young Talents Plan of China, Natural Science Foundation of China (61420106007, 61671387), and ARC grant (DE140100180). }

\balance
{\small
\bibliographystyle{ieee}
\bibliography{Referecen_Saliency}
}

\end{document}